# Exploring molecular assembly as a biosignature using mass spectrometry and machine learning

Lindsay A. Rutter, Abhishek Sharma, Ian Seet, David Obeh Alobo, An Goto, and Leroy Cronin*

School of Chemistry, University of Glasgow, University Avenue, Glasgow G12 8QQ, UK.
*Corresponding author Email: Lee.Cronin@glasgow.ac.uk

Molecular assembly offers a promising path to detect life beyond Earth, while minimizing assumptions based on terrestrial life. As mass spectrometers will be central to upcoming Solar System missions, predicting molecular assembly from their data without needing to elucidate unknown structures will be essential for unbiased life detection. An ideal agnostic biosignature must be interpretable and experimentally measurable. Here, we show that molecular assembly, a recently developed approach to measure objects that have been produced by evolution, satisfies both criteria. First, it is interpretable for life detection, as it reflects the assembly of molecules with their bonds as building blocks, in contrast to approaches that discount construction history. Second, it can be determined without structural elucidation, as it can be physically measured by mass spectrometry, a property that distinguishes it from other approaches that use structure-based information measures for molecular complexity. Whilst molecular assembly is directly measurable using mass spectrometry data, there are limits imposed by mission constraints. To address this, we developed a machine learning model that predicts molecular assembly with high accuracy, reducing error by three-fold compared to baseline models. Simulated data shows that even small instrumental inconsistencies can double model error, emphasizing the need for standardization. These results suggest that standardized mass spectrometry databases could enable accurate molecular assembly prediction, without structural elucidation, providing a proof-of-concept for future astrobiology missions.

**Introduction**

Despite advances in our exploration of the Universe, we have yet to determine whether extraterrestrial life exists. At the same time, even defining life itself remains one of the most enigmatic challenges in science[1]. Our only known example is terrestrial life, thought to have evolved from a common ancestor under conditions specific to Earth[2–4]. Traditionally, life detection missions have targeted highly specific biosignatures (chemical or physical indicators of life), while limiting the search space by relying on assumptions about life as we know it. The absence of a clear definition of life, coupled with the difficulty of interpreting the alien environments in which it might arise, has led to inconclusive outcomes.

Two notable examples of ambiguous biosignature detections within our Solar System are methane ($CH_4$) in the Martian atmosphere and phosphine ($PH_3$) in the Venusian atmosphere[5,6]. Whilst it is clear that methane can be formed biologically via methanogens (microorganisms that generate methane)[7], the biological processes thought to be involved in direct phosphine production is less clear[8]. However, these gases are also chemically simple and can arise through abiotic (non-biological) processes, such as volcanic eruptions and geothermal activity[9,10]. Complicating matters, both molecules are reactive and can be unstable in planetary atmospheres, undergoing decomposition through ultraviolet photolysis, atmospheric chemistry, and reactions with mineral surfaces[11–14]. Scientists have speculated about biological origins after detecting them at levels that purportedly exceed what known environmental processes alone can easily sustain[5,15]. Yet the dynamic balance between production and destruction of these gases makes such speculation difficult to confirm[16].

Similar ambiguities apply to biosignature candidates detected beyond our Solar System. One prominent example is dimethyl sulfide (DMS, $C_2H_6S$), tentatively identified in the atmosphere of the exoplanet K2-18b, located over 100 light-years away[17]. K2-18b is classified as a Hycean planet, a type of temperate world thought to host a subsurface ocean beneath a hydrogen-rich atmosphere[18]. On Earth, DMS can be produced by decomposition, associated with marine microorganisms, explaining why it has been suggested as a biosignature candidate[19]. However, it has also been identified in abiotic environments, such as cometary dust[20], and can be synthesized under prebiotic laboratory conditions[21]. Disentangling whether these candidate molecules originate from biological or non-biological sources requires quantifying the full range of environmental processes that could produce or destroy them, and the possibility that unfamiliar forms of life (potentially radically different from those on Earth) might be responsible for their presence. Given these many uncertainties, it remains difficult to assert that the presence of simple molecules (whether methane, phosphine, or DMS) provides clear evidence of alien life[22,23].

There is a growing interest in "agnostic" life detection strategies: methods that can identify molecular signs of life without needing an exact understanding of what that life might look like, or how it might interact with a complex, poorly characterized environment[24–26]. One alternative approach to distinguish between abiotic and biotic molecules is to assign them complexity scores[27]. These scores are fixed for each molecule and do not depend on external conditions, such as the environment in which they might be detected or the specific alien biochemistry from which they might arise. Because complexity scores are independent of such difficult-to-quantify factors, they could be considered "agnostic biosignatures". Herein we propose a new approach to life detection, which utilizes a new method to detect evidence of selection and evolution, intrinsic signatures of life, at the molecular level, see Fig. 1.

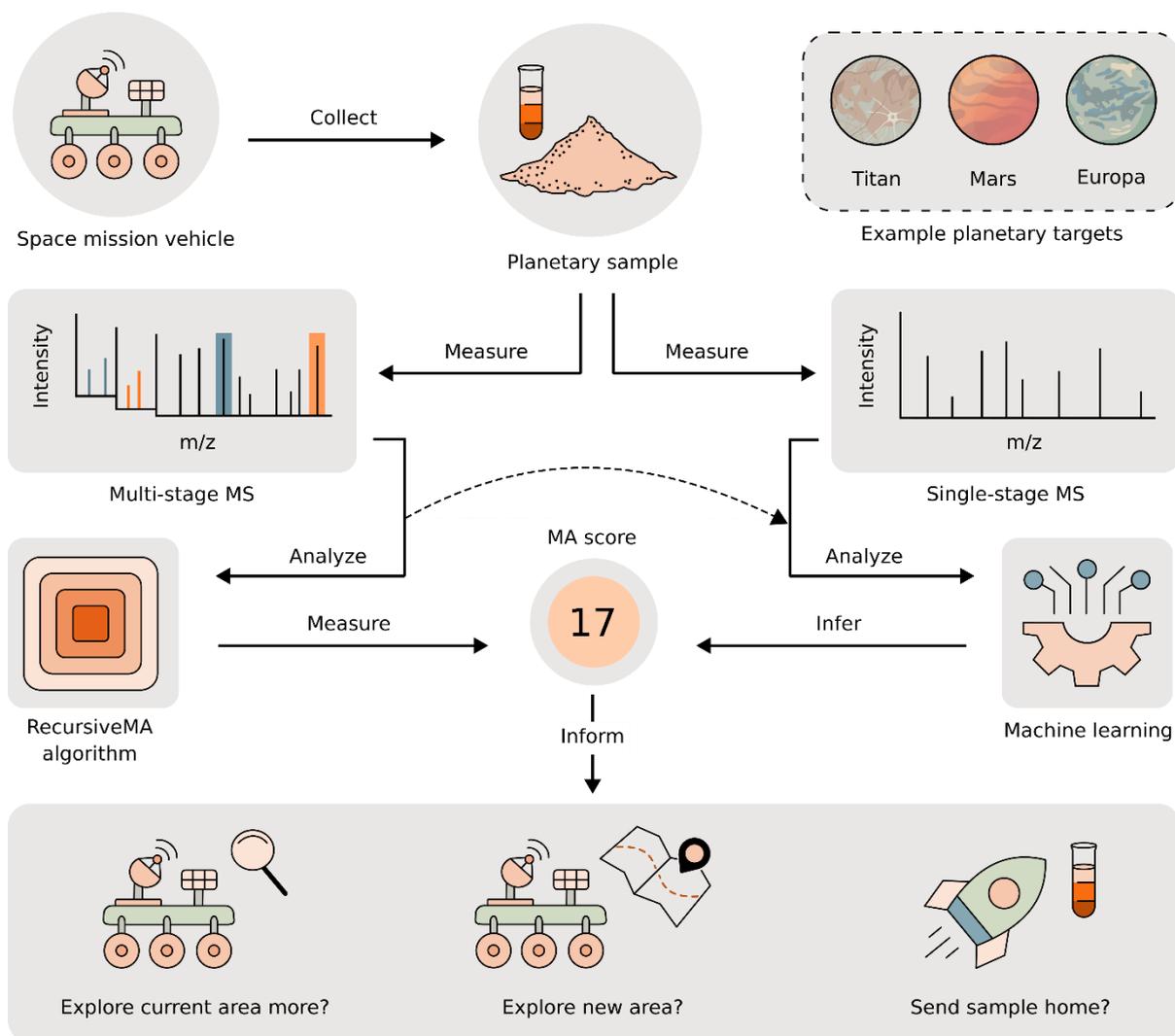

**Fig. 1: Schematic overview of deploying MA as an agnostic biosignature.** Upcoming life detection missions in the Solar System can collect samples in situ and predict MA scores directly from MS data without the need for structural elucidation, which may be difficult for unknown molecules in extraterrestrial environments. Space missions that use MS$^n$ can estimate MA scores with the RecursiveMA algorithm. In the current paper, we show that space missions that use MS$^1$ (and optionally MS$^n$) can infer MA scores with ML.

Several molecular complexity methods have been proposed, such as those by Bertz and Böttcher, which aim to quantify the structural or informational richness of a molecule[28,29]. The Bertz complexity score, one of the earliest and most influential, incorporates features from both graph theory and information theory. It estimates molecular complexity by treating the number of graph invariants of a specified type and distinct heteroatoms in a molecular graph as discrete states in Shannon's formula for information entropy. The Böttcher score, another well-established molecular complexity metric, estimates the information content in each atom's microenvironment. It does so by assessing stereogenicity, the diversity of heavy atom and bond types, and chemical equivalence[29]. Both scores use Shannon entropy-based approaches to quantify the structural diversity of a molecule, independent of how it was built.

More recently, a new approach has been developed from a theoretical framework known as assembly theory (AT), which seeks to explain how complex objects arise through processes of selection and evolution[30]. In this framework, the amount of selection (or assembly) is a function of the amount of a given object found in the environment and its assembly. To quantify the degree of selection required to construct an ensemble of molecules, AT utilizes the assembly index and the copy number of the observed objects, where the amount of selection is quantified by the assembly of the ensemble ($A$):

$$A = \sum_{i \epsilon n_i > 1}^{N} e^{a_i} \left( \frac{n_i}{N_T} \right) \tag{1}$$

where $a_i$ is the assembly index of the $i^{th}$ object, $n_i$ is its copy number of the $i^{th}$ object, $N$ is the total number of unique objects, $e$ is Euler's number, and $N_T$ is the total number of objects in the ensemble. The equation captures the delicate interplay between assembly index and copy number.

Here, the object's assembly index is defined as the length of its shortest construction pathway—that is, the minimum number of discrete steps required to build it from basic building blocks and objects in the assembly pool (i.e. intermediate objects produced in earlier steps that can be reused in subsequent ones) (Fig. 2a)[30]. When the object is a molecule, its assembly index is termed molecular assembly (MA). Because MA depends only on the shortest path, and this can be measured experimentally, this approach to molecular complexity gives a more objective approach which is less susceptible to false positives, see Fig. 2a[30]. When working with a mixture of molecules, their assembly score is termed their joint molecular assembly (JMA), which represents the shortest pathway to construct all molecules in the ensemble[30,31]. Whereas Bertz and Böttcher quantify molecular properties using entropy-based methods without regard to construction history, MA uses algorithmic approaches to compute the shortest construction pathway, capturing history-dependent complexity.

By emphasizing the history of molecular construction, MA aims to offer a deeper foundational link to life detection and origin-of-life studies. The core premise is that a molecule with a high MA (i.e. one that requires a long construction pathway) is statistically unlikely to form, especially in large quantities (i.e. high copy number), through random abiotic processes. On a barren world, the exploration of molecular space would likely be undirected, favoring the spontaneous formation of simpler molecules of lower complexity (i.e. molecules with low MA) leading to combinatorial explosion. However, under selective pressures, this exploration becomes more directed, favoring the emergence of increasingly complex molecules with longer construction pathways and greater shared historical contingencies (i.e. molecules that share more assembly pool objects with one another). The most plausible source of such pressures is biology, which evolved to encode information that constrains chemical reaction networks, guiding the reliable synthesis of highly specific and complex molecules. Consistent with this idea, experimental findings suggest that molecules with MA scores above specific thresholds may serve as reliable indicators of biology[27].

To successfully utilize MA as an agnostic biosignature, we need experimental methods to quantify the MA of molecules in an unknown sample. Many upcoming missions—including the Europa Lander, the Enceladus Orbilander, and the Dragonfly expedition to Titan—will carry mass spectrometers to analyze molecular samples directly on planetary surfaces[32]. In these remote environments, where structural elucidation is infeasible due to unfamiliar molecules or limited instrument resolution, the ability to measure molecular complexity directly from mass spectrometry (MS) data will be critical. Among existing approaches to molecular complexity, MA can be measured by a range of techniques, including multi-stage mass spectrometry (MS$^n$)[33]. In MS$^n$, a parent molecule is sequentially fragmented into smaller ions through collision-induced dissociation, and this process is repeated across multiple stages until the smallest fragments can no longer be broken down. The resulting fragmentation tree reveals the shortest pathway of assembly steps required to reconstruct the parent molecule, allowing the MA to be directly measured[33]. To operationalize this approach, the RecursiveMA algorithm was developed to successfully quantify MA scores directly from high-resolution, MS$^n$ data (Fig. 1). In contrast, scores like Bertz and Böttcher lack this physical connection to experimental data, as they rely on global, entropy-inspired properties that do not reflect stepwise construction or fragmentation, i.e. there is no physical technique that can be used to measure the complexity of the sample directly unless the structure has already been determined.

An effective agnostic biosignature must satisfy two key criteria: it must be interpretable (can distinguish between biology and non-biology), and it must be experimentally determinable. As previously outlined, MA satisfies both. First, it is interpretable: it exhibits threshold scores above which molecules are unlikely to arise abiotically, as they suggest a history of directed construction by life[27]. Second, it can be measured: this is done by applying RecursiveMA on MS$^n$ data[33]. Importantly, this does not require structural elucidation, making it highly suitable for unknown compounds in extraterrestrial settings. For these reasons, MA has been specifically proposed as a tool for agnostic life detection[27].

Considering this, an ideal strategy would be to deduce MA directly from $MS^n$ data using RecursiveMA, ensuring full mechanistic interpretability: that is, understanding exactly how the MA score is derived from the combination of fragments that reconstruct the parent molecule in the fewest steps, see Fig. 1. However, RecursiveMA cannot be used to measure MA from single-stage mass spectrometry ($MS^1$) data directly, because it requires fragment information across multiple stages[33]. Even when $MS^n$ data are available, limited fragmentation stages, poor fragmentation behavior, or limited mass resolution, may prevent accurate MA determination. In such cases, it is worth exploring whether machine learning (ML) could assist in predicting MA scores from available MS data (Fig. 1)[34]. To investigate this possibility, we evaluated the prediction of MA from $MS^1$ data acquired under conditions like those proposed for upcoming missions to Titan and Europa. These missions will use gas chromatography-mass spectrometry (GC-MS) with electron ionization (EI), which produces $MS^1$ data. Because algorithmic reconstruction of MA prediction is not possible with $MS^1$, we conducted a proof-of-concept study to assess whether ML models could learn to predict MA directly from $MS^1$ data (Fig. 1)[34].

In this study, we trained ML models on data from the NIST (National Institute of Standards and Technology) Chemistry WebBook (Standard Reference Database [SRD] 69)[35], a curated repository of EI GC-$MS^1$ data. We found that XGBoost[36], a gradient-boosted decision tree model, substantially outperforms baseline models[34]. The model also generalizes well to other $MS^1$ databases. Using simulated $MS^n$ data, we further show that model accuracy depends on consistent instrument parameters (such as ionization energy), highlighting the importance of using standardized data. Taken together, these findings suggest that when algorithmic approaches are not feasible, ML models trained on relevant, standardized MS data offer a viable alternative for in situ estimation of molecular assembly. This approach can help prioritize, in real time, which regions should be further explored, and which samples should be cached for future return to Earth.

**Rationale for selecting MA index over conventional molecular complexity scores**

In this section, we use theoretical reasoning and large-scale analysis to identify MA as the most suitable candidate for agnostic biosignature detection, based on its interpretability. We first compare how three candidate approaches (Bertz, Böttcher, and MA) scale with number of bonds ($N_B$) by plotting this information for the set of molecules used in our ML model (i.e. for all molecules with MS[1] data in the NIST SRD database) (Fig. 2b-d). We found that Bertz increases more than linearly with $N_B$ (superlinear growth), Böttcher increases approximately linearly, and MA increases less than linearly (sublinear growth) and these findings were consistent when we examined databases larger than NIST SRD, see Table 1.

| β (R²) | NIST SRD (n=32,603) | COCONUT 2.0 (n=526,968) | PubChem (n=99,129,693) |
|---|---|---|---|
| Bertz | 1.46 (0.76) | 1.34 (0.79) | 1.49 (0.85) |
| Böttcher | 0.96 (0.67) | 1.09 (0.74) | 1.00 (0.74) |
| MA | 0.74 (0.74) | 0.68 (0.64) | 0.74 (0.72) |

**Table 1: Relationship between complexity scores and molecular size.** Estimated power-law scaling between complexity scores (MA, Böttcher, and Bertz) and number of bonds ($N_B$), based on log-log regression. The scaling exponent (β) indicates whether each measure grows superlinearly (β>1), linearly (β≈1), or sublinearly (β<1) with $N_B$. R² values indicate the goodness-of-fit for the log-log linear model. The results were consistent between three databases (NIST SRD, COCONUT 2.0, and PubChem).

Bertz and Böttcher have variances that increase with increasing $N_B$, as evidenced by their funnel-shaped distributions in Fig. 2b-c. The theoretical bounds of MA scores also exhibit this feature, as shown by the two dotted red lines in Fig. 2d. However, the observed molecules do not exhibit such variances that increases with $N_B$ in their MA: with increasing $N_B$, we no longer find any molecules that adhere to the upper bound of $MA = N_B - 1$, because large molecules form in high copy numbers more readily if they reuse assembly pool objects that keep their shortest pathways of construction minimal, see Fig. 2d. These observations suggest that MA captures more underlying structures and submolecular symmetries, while Bertz and Böttcher are sensitive to total number of bonds.

We next calculated the relative mean squared error (MSE) for simple baseline models (linear regression, logarithmic regression, power law, and polynomial regression) regressed on the weight of the molecule, the number of bonds in the molecule, the mass of the maximum peak in its spectra, and the number of peaks in its spectra (Supplementary Table 1). We note that structural elucidation is required to determine the $N_B$ and molecular weight (MW) of a molecule, the latter of which is not always derivable from the $MS^1$ data, given that the parent molecule is often fragmented. For all three measures, the models regressed on features that require structural eludication (i.e. $N_B$ and MW) and predicted complexity better than features that do not require structural elucidation (i.e. number of peaks and maximum peak, Supplementary Table 1). This suggested that molecular complexity cannot be predicted well from simple baseline models applied directly to $MS^1$ data alone, and highlighted again the overall motivation for our current study (Supplementary Table 1).

It is difficult to interpret how Böttcher and Bertz scores scale; for example, it is unclear what a Bertz score of 4,000.17 means relative to scores of 40.23 or 400.09. In contrast, MA provides interpretabilities that scale: a molecule with an MA of 40 has a shortest pathway twice as long as a molecule with an MA of 20, see Fig. 2a as an example. Within the context of AT, the combinatorial spaces of physically plausible molecules in the absence of any selective bias is given by *assembly possible*, where molecules are represented by their shortest construction paths[30]. With an increase in MA, the number of combinatorial construction operations leading to all physically plausible molecules increases exponentially. As an example, the molecule with an MA of 40 has a vastly larger assembly possible space compared to the molecule with an MA of 20. The exponential expansion of assembly possible also allows for interpretation of MA differences considering single construction steps $a \rightarrow a+1$, where $a$ quantifies an intermediate construction step, as the assembly possible expands less so with an MA from 10 to 11 as compared to an MA from 20 to 21 (see Fig. 2a). In contrast, for given Böttcher and Bertz scores, it is difficult to quantify the potential combinatorial spaces of molecules as they calculate complexity based on additive and entropic properties of molecular substructures instead of path-dependent features in assembly spaces. This is because the expansion of combinatorial space of molecules cannot be interpreted intuitively with Böttcher and Bertz scores whereas the scalable interpretability of MA has been investigated across a variety of additional context, see Table 2.

|   | **Characteristic** | **MA** | **Böttcher** | **Bertz** |
|---|---|---|---|---|
| a | Dependence on molecular size | Sublinear | Linear | Superlinear |
| b | Interpretability | High, unique construction pathway | Low, hard to interpret without context | Low, hard to interpret without context |
| c | Proposed for life detection | Experimental evidence of non-life/life threshold[27] | Not currently | Not currently |
| d | Interpretable for individual and ensemble of molecules | Scores for both (MA[30] and JMA[31]) | Score for only individual molecules | Score for only individual molecules |
| e | Predictable from MS | Algorithmic, related to MS[33] | Entropic, unrelated to MS | Entropic, unrelated to MS |
| f | Computation time from known structure | High | Medium | Low |

**Table 2: Motivation for selecting MA for life detection.** Comparison of three molecular complexity approaches (MA, Böttcher, and Bertz). **a-e,** Five key advantages that identify MA as the leading agnostic biosignature candidate for space missions that utilize MS[n]. **f,** Bertz scores tend to be the fastest to compute.

The same set of molecules yield substantially different ranges across the three different approaches, with maximum MA: 24, Böttcher: 911.25, and Bertz: 2,515.02, see Fig. 2. Because each complexity measure captures different molecular attributes, the molecules ranked as low or high in complexity vary depending on the metric. Here, Bertz tends to classify linear hydrocarbon chains with simple or no branching structures as low in complexity, and molecules with dense, intricate bonding as high in complexity. In contrast, Böttcher evaluates the local microenvironment of atoms, ranking molecules with multiple stereocenters and conformational flexibility as more complex, while considering rigid and achiral molecules (such as symmetric cages or fullerenes) as less complex. Meanwhile, MA considers how many steps are required to form molecules, and hence it ranks heterogeneous molecules with few reusable substructures as high in complexity, and molecules that can recursively reuse substructures in their assembly pools as low in complexity.

It is noteworthy that, when normalized for size, polycyclic aromatic hydrocarbons (PAHs) are typically ranked as highly complex according to Bertz, but as having low complexity according to MA, see Fig. 2. PAHs are among the most abundant type of molecules in space, believed to comprise 15% of the interstellar carbon[37]. They are also widespread[38], having been detected in interstellar dust (where stars and planets form)[39], in meteorites[40], and in the atmosphere of Titan[41]. The untested "PAH world" hypothesis even proposes that PAHs facilitated the synthesis of the first RNA molecules, potentially playing a major role in the origin of life[42]. Collectively, these observations underscore how different complexity scores can result in contrasting perspectives on molecules with significant astrobiological implications.

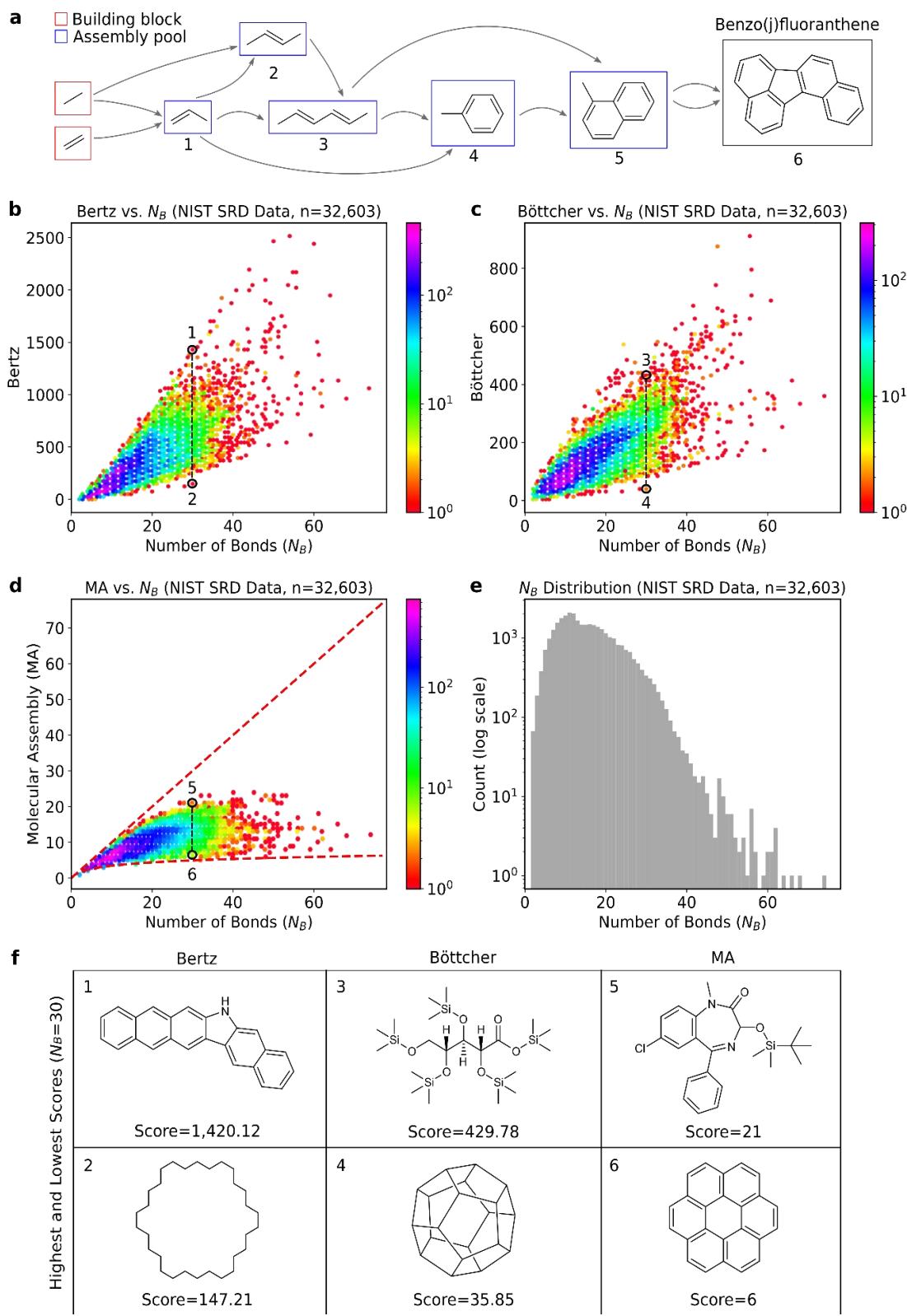

**Fig. 2: Relationship between complexity score and $N_B$. a,** Example shortest pathway construction for a molecule with MA=6. **b-d,** Scatterplots of complexity scores versus $N_B$. The two dotted red lines demonstrate mathematically derivable bounds of MA scores, with an upper bound of $MA = N_B - 1$ and a lower bound of $MA = \log_2(N_B)$. The color legends indicate the count of molecules on a logarithm scale. **e,** The distribution of $N_B$ for the 32,603 molecules with MS[1] on the NIST SRD database[35]. **f,** Highest and lowest scoring molecules with $N_B = 30$.

We also chose to focus on MA because, as far as we know, it has been specifically developed for life detection methods, following experimental evidence suggesting that MA values above certain thresholds may strongly indicate the presence of life[27]. If such thresholds exist for MA, they may be linked to more intuitive scalability and the statistical improbability of observing a large "copy number" of any given high MA molecule by random, undirected processes. This improbability arises not only from the large number of assembly steps required to construct the molecule, but also from the exponentially larger space of similarly complex molecules that could have been formed. We note that this copy number criterion is inherently satisfied in MS data, as modern instruments only detect molecules once a sufficiently large number of copies are present[43] (Table 2c).

Experimental work has suggested that biological samples have a wider distribution of MA values compared to abiotic samples, and that only biological samples produce MAs above a certain threshold (ca. 15)[27]. We investigated how well separating the COCONUT 2.0 database[44] into two groups of molecules based on MA thresholds of between 15 and 20 could similarly separate the molecules into the same two groups based on Bertz and Böttcher scores (Supplementary Figs. 2 and 3). This analysis suggests that MA thresholds that have FPRs of close to zero, based on prior experimental work, cannot be as readily translated into Bertz and Böttcher thresholds without introducing FPRs in the range of 0.17 to 0.32 (depending on the MA threshold).

The observations that reveal the relationship between Bertz and MA values for all molecules in the COCONUT 2.0 database[44] are illustrated in Fig. 3. While there is a relationship between these scores (primarily due to all complexity scores in general becoming higher with increasing molecular size), we still see significant discrepancies. For example, the three molecules shown all have the same Bertz score of around 1,000, with molecule #2 representing the average situation with an MA score of 15. However, molecule #3 has an MA of 25, which is more than four times higher than the MA score of 6 for molecule #1. This means that even though molecule #3 requires more than four times the number of assembly steps to form its shortest pathway compared to molecule #1, they still obtain the same Bertz score (see the shortest pathways for these molecules drawn out in Supplementary Figs. 3 and 4).

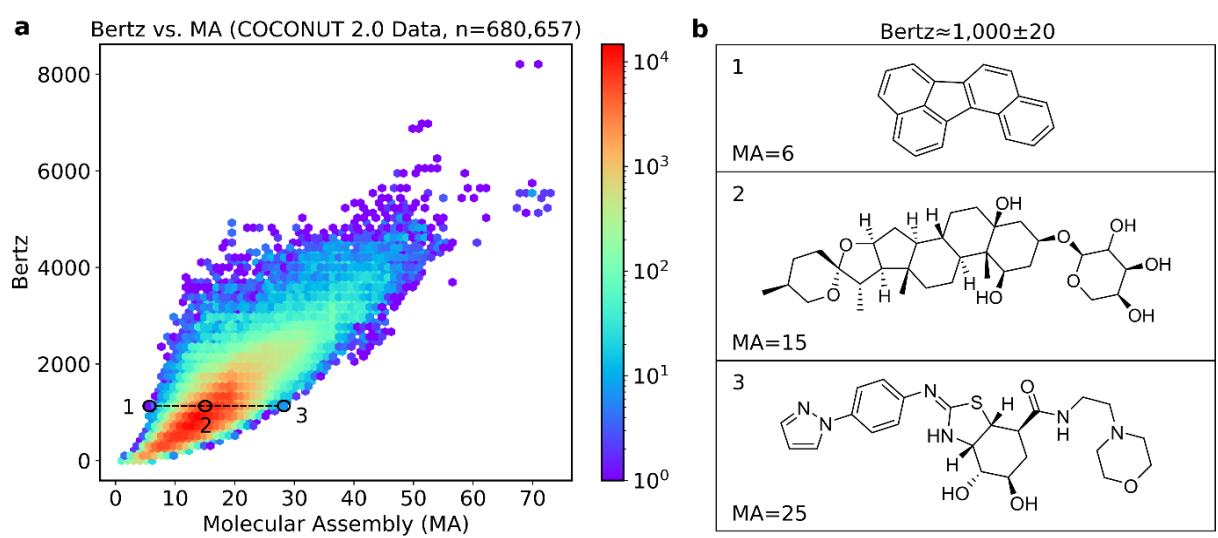

**Fig. 3: Comparing MA and Bertz scores, conditioned on constant Bertz scores. a,** Bertz versus MA scores for each molecule in the COCONUT 2.0 database[44], with the color legend indicating the count of molecules on a logarithm scale. **b,** Three example molecules with similar Bertz scores, but with either low, intermediate, or high MA scores.

Altogether, these example molecules in Fig. 3 illustrate that Bertz scores can be particularly high for periodic molecules with many bonds, even when their structural heterogeneity is limited. Molecule #1 (benzo[j]fluoranthene, B*j*F, which has been detected in meteorites[45]) has a Bertz score of around 1,000, which corresponds to an average MA of around 15, which would be high enough to warrant consideration as a biosignature.

However, B*j*F has a true MA of 6, which is much lower than what would constitute strong evidence of biological origins. This again illustrates that simple translation between the complexity scores is not possible. We showcase various other example molecules with strikingly different scores between complexity approaches in Supplementary Figs. 5-9.

**a** Dodecane

|          | 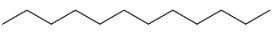 | 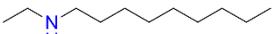 | 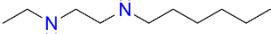 |
|----------|------|-------|--------|
| MA       | 5    | 6     | 7      |
| Böttcher | 32.00 | 83.29 | 102.58 |
| Bertz    | 56.44 | 61.40 | 64.24  |

**b** Cubane

|          | 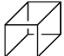 | 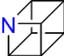 | 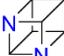 |
|----------|-------|--------|--------|
| MA       | 4     | 5      | 5      |
| Böttcher | 14.34 | 43.66  | 51.48  |
| Bertz    | 110.04 | 114.39 | 116.53 |

**c** Coronene

|          | 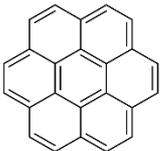 | 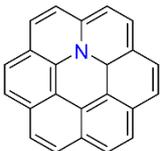 | 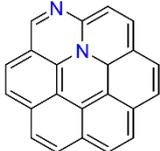 |
|----------|--------|---------|---------|
| MA       | 6      | 10      | 11      |
| Böttcher | 91.02  | 275.00  | 286.43  |
| Bertz    | 1,140.95 | 1,151.49 | 1,195.78 |

**d** Saturated Buckminsterfullerene

|          | 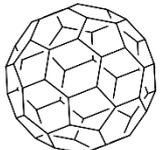 | 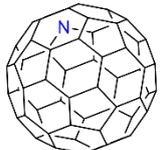 | 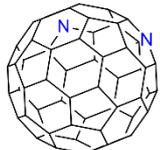 |
|----------|--------|---------|---------|
| MA       | ≤10    | ≤14     | ≤17     |
| Böttcher | 107.55 | 388.46  | 407.03  |
| Bertz    | 1,348.53 | 1,355.87 | 1,361.18 |

**Fig. 4: The effect of symmetry breakage on complexity values. a-d,** Highly symmetrical hydrocarbon molecules dodecane, cubane, coronene, and saturated buckminsterfullerene. From left to right: the highly symmetrical hydrocarbon molecule, the same molecule with one nitrogen replacing one carbon, and the same molecule with a second nitrogen replacing a second carbon. We used the saturated version for buckminsterfullerene to avoid kekulization issues that would affect MA calculations (see Supplementary Methods Section 1.5).

We highlight additional differences between the complexity values through a deliberate symmetry-breaking analysis, see Fig. 4. For this, we start with a set of highly symmetrical hydrocarbon molecules—dodecane (the largest organic molecule discovered to date on Mars)[46], cubane, coronene (present in meteorites and the interstellar medium (ISM))[47], and buckminsterfullerene (also present in meteorites and the ISM)[48,49]—and then break their symmetry by adding "defect" atoms, see Fig 4. Here, Böttcher scores increase significantly, while Bertz scores and MA increase only slightly. In this demonstration, MA and Bertz are sensitive to both full and submolecular symmetries, whereas Böttcher is sensitive to the full molecular symmetry and largely unaffected by submolecular symmetries.

To further examine the role of MA in life detection, we investigated how well the three approaches distinguished between similarly sized molecules from the PubChem and COCONUT 2.0 databases (Fig. 5)[44,50]. In a broad manner, the PubChem database[45] can serve as a proxy for what molecules are possible given the environmental constraints on planet Earth, whereas the COCONUT 2.0 database[44] can serve as a proxy for what biologically derived natural product molecules are the outcome of evolution and selection given these same constraints. Fig. 5 shows that, as expected, for both databases, both the mean and variance of Bertz and Böttcher scores generally increase as $N_B$ increases. However, Bertz scores tend to be higher for PubChem molecules compared to COCONUT 2.0 molecules of the same size, whereas Böttcher scores show the opposite trend. MA expectedly show the sublinear trend seen earlier in Fig. 2d, and higher values are seen in PubChem molecules compared to same size COCONUT 2.0 molecules.

Intriguingly, MA appears to distinguish between the databases starting from smaller $N_B$ values: for example, at $N_B = 20$, MA separates the databases such that the mean of one coincides with the standard deviation of the other (Fig. 5c). This is not seen in the cases of Bertz and Böttcher, where the means and standard deviations are not readily differentiable between the databases (Fig. 5a,b). This pattern remains even when we subset the databases to only consider molecules that strictly only contain carbon, hydrogen, nitrogen, and oxygen (Supplementary Fig. 11), and it remains (albeit in a reduced manner) when we examine the inverse set of molecules (i.e. those that contain at least one atom that is *not* carbon, hydrogen, nitrogen, and oxygen) (Supplementary Fig. 12). Altogether, this suggests that MA may detect differences between biological and non-biological origin starting from smaller sized molecules. We also examined the upper-bound of MA across both databases, and found that, for each $N_B$ value, the largest MA from PubChem was greater than or equal to the largest MA from COCONUT 2.0 (Supplementary Fig. 13)[44,50]. The largest $N_B$ value that maintained the strict upper-bound of $MA = N_B - 1$ occurred for $N_B = 14$ in the COCONUT 2.0 database, and occurred much later, at $N_B = 24$ in the PubChem database.

Upon inspecting these unusually heterogenous molecules, we found that PubChem contains pharmaceutical compounds that require synthetic intervention, thus allowing for complex molecules to occur, even at relatively smaller molecular sizes. These PubChem molecules that contain large MA values (especially above any $N_B$-matched COCONUT 2.0 molecules) can potentially be studied as "molecular technosignatures", i.e. compounds that are unlikely to occur naturally without technological intervention[51,52] (Supplementary Fig. 13). An additional reason we chose MA for this study was because algorithms like RecursiveMA can predict MA from MS$^n$ data, given that smaller peak masses of later stages add to form larger peak masses of earlier stages in recursive manners that parallel the joining operations of AT[33]. In contrast to the more abstract joining operations of molecular graph correlates, Böttcher and Bertz depend heavily on explicitly labeled molecular properties (such as bond types, atom types, valence electrons, stereocenters and their configurations, etc.)[28,29], and thus require greater structural elucidation than approaches based on features more interpretable from MS data, such as the additive properties of peak masses. As a result, as far as we know, it has not yet been shown whether an algorithm could successfully predict Böttcher and Bertz values directly from MS data (Table 2e).

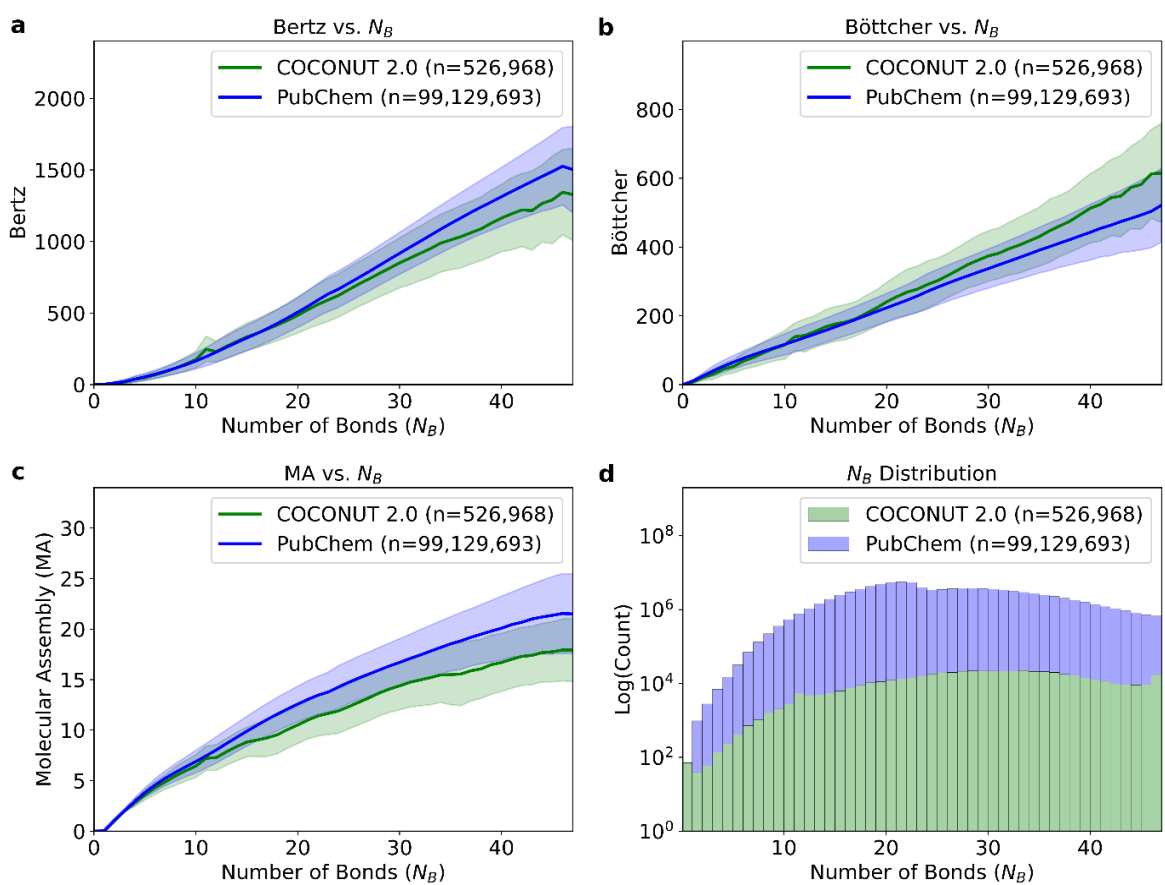

**Fig. 5: Molecular complexity growth rate per bond count between PubChem and COCONUT 2.0 molecules.** Mean (solid line) and standard deviation (shaded area) of complexity scores per bond count between PubChem molecules in blue (n ≈ 100,000,000) and COCONUT 2.0 molecules in green (n ≈ 525,000) for **a,** Bertz, **b,** Böttcher, and **c,** MA. **d,** The databases had similar distribution of bond counts in their molecules.

Altogether, we chose MA several reasons: it appears to detect underlying structure well, it provides interpretable scalability, it has shown experimental promise as a life detection tool, it provides measurements for both individual molecules and ensembles of molecules, and it can be measured from MS$^n$ data (Table 2a-e). For future life detection missions using MS$^n$, MA scores can be calculated with the RecursiveMA algorithm. This study explores whether ML can be used to predict MA scores for missions that rely on $MS^1$ data.

**Limitations of algorithmic prediction of MA from $MS^1$ data**

We briefly describe the RecursiveMA algorithm and demonstrate its intended uses and limitations. As mentioned earlier, RecursiveMA estimates MA from $MS^n$ data[33]. It accomplishes this by setting an upper limit of MA estimation based on the parent mass as a first order approximation, and then correcting this score as overlaps in child fragments are discovered[33]. In theory, complete information of all parent-child fragments in $MS^n$ data with infinite resolution should allow for the near-perfect measurement of MA. However, real-world $MS^n$ data provide only limited fragmentation information (due to neutral fragments, rearrangement reactions, limited recursions, loss of parent masses, and different bonds possessing different strengths). As a result, MA values are predicted well, but not perfectly, when using RecursiveMA in real-world $MS^n$ data[33].

While some upcoming life detection missions in our Solar System will use $MS^n$, many will use $MS^1$ due to size and power constraints, the need for speed and efficiency, and the focus on in situ analysis. However, RecursiveMA is not designed for $MS^1$ data, as such data does not include the relationship between parent and fragment ion masses across stages. In some $MS^1$ data, parent ions are also lost, which further reduces the suitability of RecursiveMA[33]. This is especially true in "hard" ionization techniques, such as electron ionization (EI) approaches, which will also be used in upcoming life detection missions in our Solar System (Supplementary Fig. 15a). These limitations motivated us to create a ML model that can predict MA scores from $MS^1$ data as a complementary approach to RecursiveMA that can predict MA scores from $MS^n$ data.

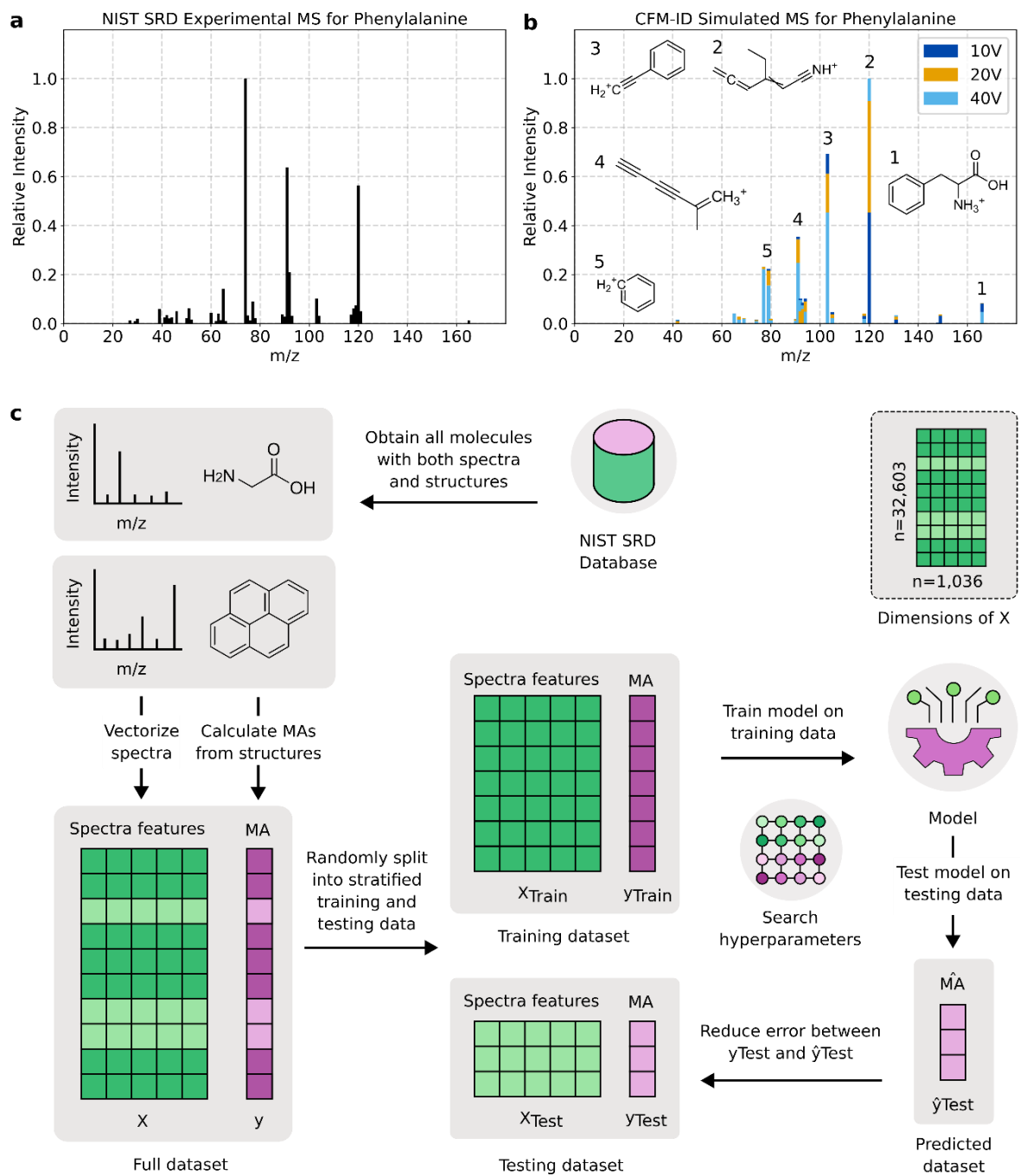

**Fig. 6: Flowchart of XGBoost model generation. a,** Experimental MS of phenylalanine from the NIST SRD database (usually MS[1] EI-GCMS 70eV). **b,** Simulated MS of phenylalanine from the CFM-ID database (MS/MS (MS[2]) ESI) at 10eV, 20eV, and 40eV. Simulated data provided fragment identification and allowed us to determine differences in m/z and intensity of peaks based on collision energy. **c,** We obtained all molecules in the NIST SRD data that contained both MS[1] and an elucidated structure. The explanatory variable was the vectorized MS[1] data, where each vector index contained the peak intensity at the corresponding m/z value, with intensities normalized between zero and unity. The response variable was the true MA scores. We randomly split our full data into training (70%) and testing (30%) data using a stratified option that ensured equal distributions of true MA scores. We trained our XGBoost model on the training data with an extensive hyperparameter grid search and selected the model that minimized the relative MSE between the true and predicted MA scores of the testing data.

**Evaluating ML prediction of MA from MS¹ data**

We assessed whether ML could allow for MA prediction from MS$^1$ data (Fig. 6a,c). For this, we trained an XGBoost[36] model on all molecules with MS$^1$ data on the NIST SRD database[35], after removing a small subset of molecules that were unsuitable for analysis (Supplementary Fig. 1). We found that our model reduced the relative MSE by three-fold (from 0.12 to 0.04) as compared to the best-performing baseline model (Supplementary Table 2).

We observed that molecules with low MAs tended to be overpredicted and molecules with high MAs tended to be underpredicted (Fig. 7). The molecules with low MAs were likely overpredicted due to the strict lower bound of zero for possible MA. Importantly, the underprediction of high MA molecules ensured that our model remained conservative, mitigating the risk of falsely categorizing molecules as stronger life indicators than they might truly represent. We also performed various quality checks and confirmed that our XGBoost model demonstrated the ideal learning curve (Supplementary Fig. 14), implying that our model was neither under- nor over-fitting to the training data. Next, we evaluated whether our well-performing XGBoost model, trained on the NIST SRD database[35], could generalize to other datasets. For this, we obtained all EI-B MS$^1$ data from the MassBank repository[53] and removed any molecules that had already been trained in the NIST SRD database. Then, we ran our XGBoost model from the NIST SRD data on the MassBank EI-B MS$^1$ molecules, resulting in a relative MSE of 0.07. This still demonstrated an improvement over the best power law model fitted on molecular weight (0.12) (Supplementary Table 2). We noticed that MassBank allows users to fill out more than 100 metadata fields for their uploaded samples, with many fields encompassing spectral information, such as instrument type, run time, and collision energy. While our XGBoost model confirmed that ML can predict MA reasonably well from MS, we wished to investigate how these MS instrumental parameters affect the accuracy of this approach. In the next section, we investigate this question in detail by using simulated MS data.

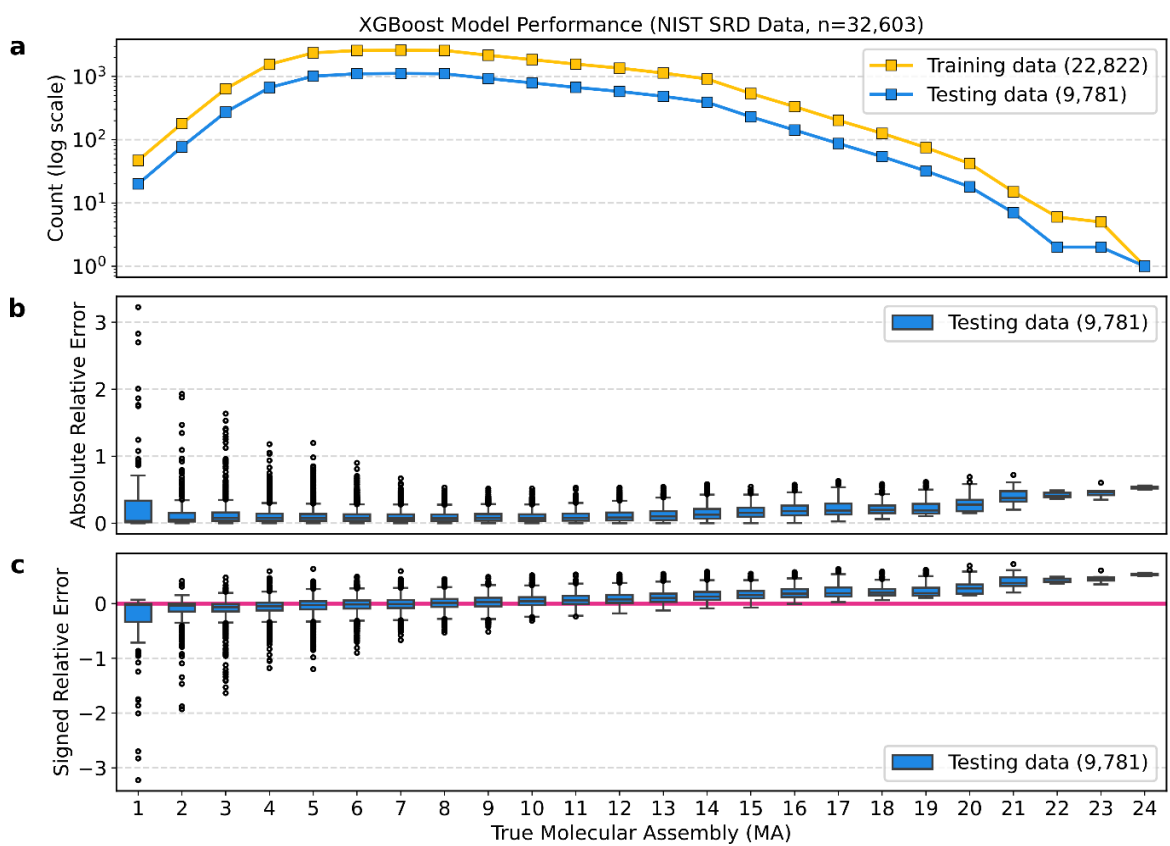

**Fig. 7: XGBoost model performance on the NIST SRD data. a,** Count of molecules per MA bin for the training and testing data, **b,** absolute relative error (calculated as $|y - \hat{y}|/y$), and **c,** signed relative error (calculated as $(y - \hat{y})/y$)), where $y$ equals the true MA score and $\hat{y}$ equals the predicted MA score. Comparing **a** and **b**, we see an anti-correlation between the absolute relative errors and the count of molecules for each MA score in the data. As **c** shows, low MA molecules tended to be over-predicted (leading to a negative signed prediction error) and high MA molecules tended to be under-predicted (leading to a positive signed prediction error). This trend revealed that our model was conservative in mitigating false positive assignment of life detection (i.e. was unlikely to overestimate the MA scores of high MA molecules).

**Determining instrument effects on ML error using simulated MS**

We generated theoretical MS data for the same molecules from the NIST SRD database using the Competitive Fragment Modeling IDentification (CFM-ID) version 4 software[54]. For each molecule, we generated three Electrospray Ionization (ESI) Quadrupole Time of Flight (QToF) [M+H]+ MS$^2$ (MS/MS) data, with low (10eV), medium (20eV), and high (40eV) collision energy levels. The CFM-ID software was able to produce spectra for all the NIST SRD molecules used in our main study, except for fourteen molecules, which we therefore excluded. We trained three XGBoost models (one for each energy level) with the same approach we used for the experimental NIST SRD dataset (Table 4, Fig. 6). Compared to the experimental data, the simulated data further reduced the relative MSE to values of 0.033 (for the 10eV-only data), 0.031 (for the 20eV-only data), and 0.031 (for the 40eV-only data), yielding a four-fold (0.12 to 0.03) reduction in our error metric compared to the best-performing baseline model (diagonal cells in Table 4, Supplementary Table 2). It is likely that simulated MS data mitigates the inherent noise in experimental data, enhancing the predictability of MA scores, although it is not entirely a fair comparison as these two datasets had other differences (the NIST SRD data was $MS^1$ that used EI, whereas the CFM-ID data was $MS^n$ that used ESI).

We also trained an XGBoost model on split-energy data, where one-third of the MS data was from each of the three energy levels respectively. Our split-energy model resulted in a relative MSE of 0.035, which was larger than any of the single-energy models (diagonal cells in Table 4). These results verified a decrease in the accurate prediction of MA when the data used non-consistent instrument parameters. These observations were expected as we had already confirmed that different energy levels capture different information about the MS fragments (Fig. 6d, Supplementary Figs. 15-17).

If it is straightforward to measure MA with MS, then it would hypothetically be more powerful to combine information across various energy levels to simultaneously capture as many different subsets of MS fragments. To investigate this, we trained an XGBoost model on integrated-energy data, where each sample contained concatenated spectral data from all three energy levels (10eV, 20eV, and 40eV), providing a combined representation. As expected, our integrated-energy model performed the best of the five models, achieving the lowest relative

MSE of 0.029 (diagonal cells in Table 4, Supplementary Fig. 18). The XGBoost model trained on each of our five models (diagonal cells in Table 4) was then evaluated with the test sets from the other four models (non-diagonal cells in Table 4). For each XGBoost model trained on a single-energy level, the relative MSE increased when the testing set came from one of the other two energy levels or the mixed-energy level, with an almost doubling of error when the model trained on the 40eV data was tested on the 10eV data (0.031 to 0.060) (Table 4). For our XGBoost model trained on the integrated-energy level, we observed a more than 2.3-fold increase in error when tested on the 10eV data (0.029 to 0.067) (Table 4). These findings highlight the importance of consistent instrumental parameters between the training and testing data for optimal prediction of MA from MS datasets.

Even though our model performs much better than baseline models, we conducted several post-hoc analyses to determine the underlying causes of over- and under-prediction of MA in our XGBoost models (see Supplementary Materials Section 4). We found that heavier molecules that break into an unusually large number of fragments (such as molecules with long hydrocarbon chains) tend to increase the chance of an overprediction, while heavier molecules that do not break at all or only break into an unusually small number of fragments (such as molecules with fused benzene rings and steric hindrance) tend to increase the chance of an underprediction (Supplementary Figs. 19-21). In fact, some of the molecules that were unusually underpredicted in our model simply flew through the MS instrument without breaking into any fragments at all, resulting in only a single peak (Supplementary Fig. 21).

Hence, it seems that our XGBoost model itself does a suitable job with learning how to predict MA scores from MS data, as corroborated by our smooth and convergent learning curves (Supplementary Fig. 14). Instead, the current limitations are not so much due to the model itself, but due to MS instrumentation constraints. Given that our model performed best with the integrated-energy data in Table 4, we believe it is likely that multimodal approaches could improve our model even further. For example, a multimodal model that incorporated MS data from energy levels even harsher than 40eV could potentially allow for the fragmentation of molecules that are otherwise too stubborn to break at all in lower energy levels, decreasing their tendency for underprediction (Supplementary Fig. 21). Altogether, it is likely our XGBoost model would effectively learn and benefit from such multi-modal MS information.

**Conclusions**

In this work, we presented a proof-of-concept demonstrating the potential use of ML to predict molecular complexity as a biosignature from in situ mass spectrometers during future space missions. Although the application of ML in life detection remains a nascent endeavor, it holds significant promise[55]. Our work builds upon recent advancements in the field, complementing ongoing efforts that underscore the growing utility of ML in the search for alien life[56,57]. Additionally, we used simulated data to highlight how instrumental parameters in MS can influence fingerprint reproducibility, potentially diminishing the effectiveness of training sets. Our study emphasizes the need for well-standardized, well-documented MS databases to optimize their use in ML applications.

Our proof-of-concept considered the MA values of individual molecules. Implementation of this metrology on other planets would require chromatographic separation of individual molecules before fragmentation in the mass spectrometer. Nonetheless, it could be of interest to conduct a similar proof-of-concept that considers the JMA indexes of ensembles of molecules. In future studies, it may also be valuable to infer MA from nuclear magnetic resonance (NMR) data using ML[33]. If it is true that NMR approaches involve less variability in instrumental parameters compared to MS[58], then this could lead to better standardized data and, consequently, improved model performance. Although NMR is less likely to be deployed in upcoming space missions due to its size, weight, and power consumption, such a project could still provide meaningful insights. For example, it could be useful for predicting MA as a biosignature in extraterrestrial samples deliberately returned to Earth[59], or in extraterrestrial samples that arrive naturally on Earth, such as the ALH84001 meteorite[60]. Inferring MA from infrared (IR) data using ML is another area of interest, especially in the case of investigating biosignatures in exoplanetary atmospheres. In some scenarios, it could be worthwhile to employ multimodal approaches that combine information from MS, NMR, and/or IR data to further improve the predictive power of molecular complexity when structural elucidation is difficult or impossible[33,61].

**Methods**

For the experimental data, we used all molecules with MS[1] data on the NIST Chemistry WebBook (SRD 69)[35]. We generated their corresponding theoretical MS$^n$ data using the CFM-ID version 4 software[54], with low (10eV), medium (20eV), and high (40eV) energy levels. Our XGBoost model[36] was developed with the Python package xgboost using a loss function of relative MSE, calculated as $\frac{1}{n}\sum_{i=1}^{n}((\hat{y}_i - y_i)/y_i)^2$. We calculated MA scores using the assemblyCPP version 5 software (which is a faster C++ implementation of AssemblyGo algorithm), Böttcher scores using code from the Forli Lab at Scripps Research (https://forlilab.org/code/bottcher/) as $\sum_i d_i e_i s_i \log_2(V_i b_i) - \frac{1}{2}\sum_j d_j e_j s_j \log_2(V_j b_j)$, and Bertz scores using the BertzCT function in the RDKit package[62]. Complete details of all methods are described in the Supplementary Information.

## Data availability

All relevant data is/will be uploaded to Zenodo and larger datasets will be available on reasonable request from the corresponding author (Lee.Cronin@glasgow.ac.uk).

## Code availability

All the codes used to perform analysis and generate figures in the manuscript and Supplementary Information are available at: https://github.com/croningp/ms2mawml.

## Conflict of interest

The authors declare no competing interests.

## Author Contributions

L.C. conceived the concept, and then the project, with L.A.R., A.S., and I.S. contributing ideas. L.A.R. performed all analyses, trained the models, and generated all figures, with A.S., I.S., and A.G. providing critical input. A.G. and I.S. contributed and guided the development of models, analyses, interpretations, and evaluations for ML. D.O.A., L.A.R., and A.S. computed PubChem and COCONUT 2.0 calculations. L.A.R. prepared the initial draft of the manuscript, with significant guidance from A.S. and L.C. All authors revised and reviewed the manuscript.

## Acknowledgements

We acknowledge financial support from the John Templeton Foundation (grant nos. 61184 and 62231), the Engineering and Physical Sciences Research Council (EPSRC) (grant nos. EP/L023652/1, EP/R01308X/1, EP/S019472/1 and EP/P00153X/1), the Breakthrough Prize Foundation and NASA (Agnostic Biosignatures award no. 80NSSC18K1140), MINECO (project CTQ2017-87392-P), and the European Research Council (ERC) (project 670467

SMART-POM). We thank Keith Y. Patarroyo for providing shortest pathway visualization tools. Valuable discussions with Amit Kahana, Louie Slocombe, Keith Y. Patarroyo, S. Hessam M. Mehr, Anne Schumacher, Stuart Marshall, and Michael Jirasek helped inform aspects of this work. We are grateful to Maria Diana Castro Spencer, Jennifer S. Mathieson, Mary Wong, and Jim McIver for providing support with resources that contributed to this work. We thank members of the astrobiology team from the 2022 Frontier Development Lab (FDL) for early work that supported this study.